\documentclass[letterpaper]{article} 
\usepackage{aaai2026}
\usepackage{times}  
\usepackage{helvet}  
\usepackage{courier}  
\usepackage[hyphens]{url}  
\usepackage{graphicx} 
\usepackage{natbib}  
\usepackage{caption} 
\usepackage{algorithm}
\usepackage{algorithmic}

\usepackage{newfloat}
\usepackage{listings}
\DeclareCaptionStyle{ruled}{labelfont=normalfont,labelsep=colon,strut=off} 
\floatstyle{ruled}
\newfloat{listing}{tb}{lst}{}
\floatname{listing}{Listing}
\usepackage{booktabs}
\usepackage{multirow}
\usepackage{amssymb}
\usepackage{mathtools}
\usepackage{amsthm}
\usepackage{cleveref}
\makeatother
\title{Credal Ensemble Distillation for Uncertainty Quantification}
\author{
    Kaizheng Wang,\textsuperscript{\rm 1,2}\thanks{Corresponding author.} Fabio Cuzzolin,\textsuperscript{\rm 3} David Moens,\textsuperscript{\rm 2,4} Hans Hallez\textsuperscript{\rm 1}
}
\affiliations{
    \textsuperscript{\rm 1}Department of Computer Science, KU Leuven, Belgium, \textsuperscript{\rm 2}Flanders Make@KU Leuven, Belgium\\
    \textsuperscript{\rm 3}School of Engineering, Computing and Mathematics, Oxford Brookes University, U.K.\\
    \textsuperscript{\rm 4}Department of Mechanical Engineering, KU Leuven, Belgium\\
    \{kaizheng.wang, david.moens, hans.hallez\}@kuleuven.be, fabio.cuzzolin@brookes.ac.uk.
}

\begin{document}

\maketitle

\begin{abstract}
Deep ensembles (DE) have emerged as a powerful approach for quantifying predictive uncertainty and distinguishing its aleatoric and epistemic components, thereby enhancing model robustness and reliability. However, their high computational and memory costs during inference pose significant challenges for wide practical deployment. To overcome this issue, we propose \emph{credal ensemble distillation} (CED), a novel framework that compresses a DE into a single model, \emph{CREDIT}, for classification tasks. Instead of a single softmax probability distribution, CREDIT predicts class-wise probability intervals that define a credal set, a convex set of probability distributions, for uncertainty quantification. Empirical results on out-of-distribution detection benchmarks demonstrate that CED achieves superior or comparable uncertainty estimation compared to several existing baselines, while substantially reducing inference overhead compared to DE.
\end{abstract}


\section{Introduction}
\begin{figure*}[t]
\centering
\includegraphics[width=0.95\textwidth]{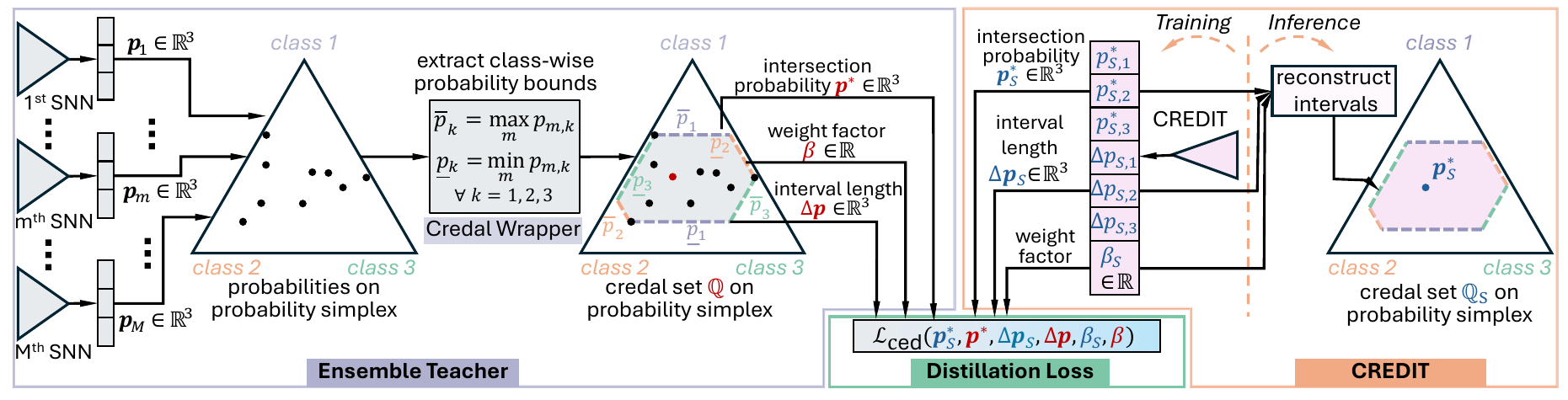}
\caption{CED framework for three-class classification ($C \!=\!3$). Given an ensemble teacher composed of $M$ SNNs, the predicted probabilities can generate class-wise probability bounds via a credal wrapper (see Sec. \ref{subsec: CredalWrapper}). These intervals form a credal set for UQ, from which a unique intersection probability is extracted for class prediction. As described in Sec. \ref{subsec: CredalStudent}, the proposed credal student is designed to output a vector $\boldsymbol{v}\!:=\!(\boldsymbol{p}_S^{*}\!\in\!\mathbb{R}^C \!, {\Delta\boldsymbol{p}}_S \!\in\!\mathbb{R}^C \!, \beta_S \!\in\!\mathbb{R})$, each component representing the intersection probability, the interval length vector, and the weight factor, respectively. The student is trained using a novel distillation loss introduced in Sec. \ref{subsec: DistillationStrategy}. At inference time, $\boldsymbol{p}_S^{*}$ is employed for class prediction, while $\boldsymbol{v}$ can recover a credal set $\mathbb{Q}_S$ for UQ.}
\label{Fig: MethodOverview}
\end{figure*}
Uncertainty quantification (UQ) in neural networks (NNs) has gained increasing attention, with two primary types of uncertainty distinguished: aleatoric uncertainty (AU), which stems from the inherent randomness in the data generation process and models the stochasticity in the output given an input (i.e., via a conditional distribution $p(\text{output}|\text{input})$); and epistemic uncertainty (EU), which is caused by a lack of evidence and reflects the model’s imprecise knowledge of the true conditional distribution  \citep{hullermeier2021aleatoric,hullermeier2022quantification,wang2024CredalEnsembles,WangReview}.
The effective estimation and differentiation of AU and EU can improve a model's trustworthiness and robustness \citep{senge2014reliable, kendall2017, Sale2023VolumeCredal, manchingal2025randomset}. For example, proper EU estimates can help avoid misclassifying ambiguous in-distribution (ID) samples as out-of-distribution (OOD), since their ambiguity does not necessarily correspond to regions of high EU within the ID distribution \citep{mukhoti2023deep,wang2025credalwrapper}. 

To quantify both AU and EU, recent studies propose training NNs to predict a \emph{second-order representation}, capable of expressing the uncertainty about a prediction's uncertainty itself \citep{malinin2019ensemble,hullermeier2021aleatoric,caprio2024credal,wang2024CredalEnsembles}. Bayesian neural networks (BNNs) \citep{blundell2015weight, gal2016dropout, krueger2017Bayesian, mobiny2021dropconnect}, in particular, learn posterior distributions over their weights and enable predictions in the form of \emph{second-order distributions} \citep{caprio2024credal}. However, BNNs generally face significant challenges in scalability to large datasets and complex architectures due to high computational demands \citep{mukhoti2023deep}. Their performance is also sensitive to the choice of prior, likelihood, and training objectives \citep{henning2021bayesian, knoblauch2022optimization}.

Alternative to BNNs, deep ensembles (DE), which combine multiple standard neural networks (SNNs) to predict \emph{a finite set of distributions} \citep{lakshminarayanan2017simple}, have been treated as a strong UQ baseline \citep{ovadia2019can, gustafsson2020evaluating, abe2022deep, mucsanyi2024benchmarking}. Nevertheless, a key limitation of DEs is their substantial demand for memory and computational resources. To this end, \emph{ensemble distillation} (ED) has become a popular way of significantly reducing inference costs \citep{hinton2015distilling,lin2020ensemble}, by distilling a DE into an SNN that approximates the mean of DE's predictive distributions. However, one drawback of ED is that the distilled SNN only generates a single predictive distribution, limiting its ability to quantify AU. This is because this single distribution captures randomness in the mapping between the input and output while assuming precise knowledge of this dependency \citep{hullermeier2022quantification}. 

To address this problem, \emph{ensemble distribution distillation} (EDD) \citep{malinin2019ensemble} has been proposed to distill a DE into a single model outputting a \emph{Dirichlet distribution} as the second-order prediction. Yet, one practical challenge in EDD and other Dirichlet-based methods (DBMs) \citep{malinin2018predictive, malinin2019reverse,charpentier2020posterior} is the absence of ground-truth Dirichlet labels for training. In addition, DBMs have recently faced criticism for departing from the theoretical tenets of epistemic uncertainty \citep{ulmer2023prior}, and failing to provide a meaningful quantitative interpretation of EU \citep{juergensepistemic}.

In an alternative approach,
\emph{credal sets}, i.e., convex sets of probability distributions \citep{levi1980enterprise}, have been employed for UQ in a broader machine learning context \citep{zaffalon2002naive, corani2008learning, corani2012bayesian, maua2017credal}. This method has recently garnered renewed attention in deep learning. Recent advancements include modeling both NN weights and outputs as credal sets \citep{caprio2024credal}, deriving credal set predictions from outputted probability intervals \citep{wang2024CredalEnsembles,wang2025creinns}, and wrapping the predictive probabilities of BNNs and DE as a credal set \citep{wang2025credalwrapper}, to name a few. Although these credal predictors offer improved UQ compared to BNN and DE baselines, they generally demand even greater computational resources for inference.

In this context, a research question arises: \textit{Can a single NN predicting a credal set as a second-order representation be distilled from a DE, which is capable of improving the UQ performance of existing distillation frameworks?}

\textbf{Novelty and Contributions} In response, this paper proposes an innovative distillation framework, termed as \emph{credal ensemble distillation} (CED), to distill a DE teacher into a single model, called \emph{CREDIT}. The distilled CREDIT can predict class-wise probability intervals \citep{probability_interval_1994} that can define a credal set.

As illustrated in Figure \ref{Fig: MethodOverview}, given multiple predictive softmax outputs from a DE teacher, the proposed CED framework first applies the credal wrapper method in \citep{wang2025credalwrapper} to construct class-wise probability intervals. From these intervals, an \emph{intersection probability} \citep{cuzzolin2022intersection} is computed—a single probability vector used for class prediction—based on the interval lengths and a scale weight factor. For a classification task involving $C$ classes, the proposed CREDIT modifies the final classification layer of a standard neural network architecture to output a vector in $\mathbb{R}^{2C+1}$. This vector consists of the predicted intersection probability in $\mathbb{R}^C$, the corresponding probability interval lengths in $\mathbb{R}^C$, and a scalar scale weight. CREDIT is trained using label information distilled from the DE teacher via a novel distillation loss. At inference time, the intersection probability is used for class prediction, while the full output vector enables the reconstruction of a credal set for uncertainty quantification.

While our work employs a recent credal wrapper \citep{wang2025credalwrapper} to extract credal information from ensembles, we emphasize that \emph{credal ensemble distillation} is a \emph{novel task} that, to our knowledge, has not been previously explored. Our objective is to design a novel and simple single model capable of learning and retaining the credal information from an ensemble, while improving the UQ performance of the ensemble and existing distillation approaches.

Empirical experiments on several OOD detection benchmarks, including different dataset pairs and various network architectures, demonstrate that CED achieves superior or comparable uncertainty estimation compared to several existing ED, EDD, and DE baselines, while substantially reducing inference overhead compared to DE.

The remainder of this paper is organized as follows. Sec. \ref{sec: background} extends the background discussion. Sec. \ref{sec: Methodology} introduces the full details of our CED. Sec. \ref{sec: Experiment} describes the experimental validations. Sec. \ref{Sec: conclude} summarizes our conclusion and future work. 

\section{Background}
\label{sec: background}
This section discusses UQ in deep ensembles and different distillation frameworks.

\textbf{Deep Ensembles (DE)\ }
DE generates an averaged prediction from a set of $M$ individually-trained standard neural networks (SNNs), as follows: 
\begin{equation}
	\textstyle\tilde{\boldsymbol{p}}\!=\!\frac{1}{M}\!\sum\nolimits_{m=1}^{M}\!\text{SNN}_{m}(\boldsymbol{x} )\!=\!\frac{1}{M}\!\sum\nolimits_{m=1}^{M}\!\boldsymbol{p}_m,
	\label{Eq: predDE}
\end{equation}
where $\boldsymbol{p}_m$ represents the single predictive probability vector from the $m$-th SNN. In this context, the total uncertainty (TU) and  the aleatoric uncertainty (AU) in DE can be approximately quantified by the Shannon entropy of the averaged prediction (denoted as $H(\tilde{\boldsymbol{p}})$) and by averaging the Shannon entropy of each sampled prediction ($\tilde{H}(\boldsymbol{p})$) \citep{hullermeier2021aleatoric}, respectively, as follows:
\begin{equation}
\begin{aligned}
&\textstyle H(\tilde{\boldsymbol{p}})\!=\!-\sum\nolimits_{k=1}^{C}\!\tilde{p}_k\log{\tilde{p}_k} \\ 
&\textstyle \tilde{H}(\boldsymbol{p})\!=\!-\!\frac{1}{M}\!\sum\nolimits_{m=1}^{M}\!\sum\nolimits_{k=1}^{C}\!p_{m,k}\log{p_{m,k}}
\label{Eq: AU_BNN}
\end{aligned}.
\end{equation}
Here, $\tilde{p}_k$ and $p_{m,k}$ are the $k$-th elements of the probability vectors $\tilde{\boldsymbol{p}}$ and $\boldsymbol{p}_m$, respectively. The level of epistemic uncertainty (EU) can be estimated as 
$H(\tilde{\boldsymbol{p}})\!-\!\tilde{H}(\boldsymbol{p})$ \citep{depeweg2018decomposition}, which can be interpreted as an approximation of mutual information (MI) \citep{hullermeier2022quantification,hullermeier2021aleatoric}.

\textbf{Ensemble Distillation (ED)\ }
Given a DE of $M$ trained SNNs, a single model, ED-Net, can be trained within the ED framework by minimizing the cross-entropy between its predictive categorical distribution ($\boldsymbol{p}$) and the soft label ($\tilde{\boldsymbol{p}}$) obtained by averaging the ensemble's output distributions in \cref{Eq: predDE}, as follows \citep{hinton2015distilling}: 
\begin{equation}
     {N}^{-1}\!\textstyle\sum\nolimits_{n=1}^{N}\textstyle\big(-\!\sum\nolimits_{k=1}^{C}\tilde{p}_k^n\log{{p}_k^n}\big)
	\label{Eq: EDLoss}.
\end{equation}
Here, the superscript $n$ is the index of the $N$ training samples. ED-Net can merely measure AU by calculating the classical Shannon entropy as $\textstyle H({\boldsymbol{p}})\!=\!-\!\sum\nolimits_{k=1}^{C}\!{p}_k\log{{p}_k}$.

\textbf{Ensemble Distribution Distillation (EDD)\ } 
In contrast to SNNs predicting a single softmax probability, the EDD-Net model distilled within the EDD framework generates the parameter vector (denoted as $\boldsymbol{\alpha}\!=\!\mathbb{R}_{+}^{C}$) of a Dirichlet distribution 
$\boldsymbol{q}\!:=\!\text{Dir}(\boldsymbol{\alpha})$. Let $\boldsymbol{z}_{\text{edd}}\!\in\!\mathbb{R}^{C}$ be the output logits of EDD-Net; both $\boldsymbol{\alpha}$ and the expected categorical distribution (denoted as $\boldsymbol{\pi}$) under this Dirichlet prior can be calculated as follows \citep{malinin2019ensemble}:
\begin{equation}
	\begin{aligned}
		&\textstyle\boldsymbol{\alpha}\!=\!\operatorname*{exp}(\boldsymbol{z}_{\text{edd}})\\
		&\textstyle\boldsymbol{\pi}\!=\!\frac{\boldsymbol{\alpha}}{\sum_{k=1}^{C}\alpha_k}\!=\!\frac{\operatorname*{exp}(\boldsymbol{z}_{\text{edd}})}{\sum_{k=1}^{C}\operatorname*{exp}({z}_{\text{edd}_k})}\!=\!\operatorname*{softmax}(\boldsymbol{z}_{\text{edd}})
		\label{Eq: predEDD}
	\end{aligned}.
\end{equation}
$\boldsymbol{\pi}$ is employed for making class prediction. TU and AU can be measured by the Shannon entropy of $\boldsymbol{\pi}$ and the expected entropy of a categorical distribution sampled from the $\text{Dir}(\boldsymbol{\alpha})$, respectively, as follows:
\begin{equation}
	\begin{aligned}
		&\textstyle H(\boldsymbol{\pi})\!=\!-\!\sum\nolimits_{k=1}^{C}\!{\pi}_k\log{{\pi}_k}\\
		&\textstyle\mathbb{E}_{\boldsymbol{p}\sim\text{Dir}(\boldsymbol{\alpha})}(H(\boldsymbol{p}))\!=\!-\!\sum\nolimits_{k=1}^{C}\!\frac{\alpha_k}{\alpha_0}\!\big(\psi(\alpha_k\!+\!1)\!-\!\psi(\alpha_0\!+\!1)\big)
		\label{Eq: uqEDD}
	\end{aligned}.
\end{equation}
Here $\alpha_0\!=\!\sum_{k=1}^{C}\alpha_k$, while $\psi$ denotes the digamma function. EU is calculated as $H(\boldsymbol{\pi})\!-\!\mathbb{E}_{\boldsymbol{p}\sim\text{Dir}(\boldsymbol{\alpha})}(H(\boldsymbol{p}))$ \citep{malinin2019ensemble}. 

\section{Methodology}
\label{sec: Methodology}
This section details our \emph{credal ensemble distillation} (CED) framework. As shown in Figure \ref{Fig: MethodOverview}, a DE teacher comprising $M$ SNNs generates class-wise probability intervals using a credal wrapper. These intervals define a credal set for UQ, from which a unique intersection probability is derived for class prediction (see Sec. \ref{subsec: CredalWrapper}). To distill a credal predictor from the DE teacher, we introduce an innovative credal student, termed as CREDIT (Sec. \ref{subsec: CredalStudent}), capable of predicting the intersection probability, the interval length vector, and a weight factor for collectively reconstructing the probability interval systems. Sec. \ref{subsec: DistillationStrategy} proposes our distillation strategy for effectively training the credal student.
\subsection{Credal Wrapper for Ensemble Teacher}
\label{subsec: CredalWrapper} 
A credal wrapper method has been recently proposed to enhance the UQ capabilities of DE by generating a credal set from DE's $M$ predicted probabilities \citep{wang2025credalwrapper}. Specifically, a set of probability intervals over $C$ classes, denoted as $[\underline{\boldsymbol{p}}, \overline{\boldsymbol{p}}]\!:=\!{\{[{\underline{p}}_k, {\overline{p}}_k]\}}_{k=1}^{C}$, can be computed from
\begin{equation}
	{\overline{p}}_k\!=\! \operatorname*{max}_{m=1,..,M}{p_{m,k}}, \quad {\underline{p}}_k\!=\! \operatorname*{min}_{m=1,..,M}{p_{m,k}}
	\label{Eq: ExtractPI},
\end{equation}
where $p_{m,k}$ is the $k$-th class value of the $m$-th predicted probability $\boldsymbol{p}_m$ from DE. These intervals can determine a valid credal set $\mathbb{Q}$ \citep{probability_interval_1994} as
\begin{equation}
\textstyle \mathbb{Q} \!=\! \{\boldsymbol{p} \!\mid\! p_k \!\in\! [{\underline{p}}_k, {\overline{p}}_k] \ \forall k\}
\label{Eq: CredalPIs},
\end{equation}
while satisfying 
\begin{equation}
\textstyle \sum\nolimits_{k=1}^{C}\! {\underline{p}}_k \!\leq 1 \! \leq \! \textstyle \sum\nolimits_{k=1}^{C}\!{\overline{p}}_k
\label{Eq: CredalPIsCondition}.
\end{equation}
As a result, $\mathbb{Q}$ consists of a convex set of valid (normalized) probability vectors $\boldsymbol{p}$, whose any $k$-th probability value $p_k$ is constrained by the probability interval $[{\underline{p}}_k, {\overline{p}}_k]$.

To derive a unique class prediction from the credal set defined in \cref{Eq: CredalPIs}, the credal wrapper \citep{wang2025credalwrapper} computes a normalized \emph{intersection probability}, $\boldsymbol{p}^*$, under the assumption of equal trust in the probability intervals across all classes \citep{cuzzolin2022intersection}.  Mathematically, the $k$-th element of $\boldsymbol{p}^*$ is obtained from
\begin{equation}
\textstyle p^*_k \!=\! {\underline{p}}_k + \beta({\overline{p}}_k\!-\!{\underline{p}}_k)
\label{Eq: intersection},
\end{equation}
where the weight factor $\beta\!\in\![0, 1]$ can be computed as
\begin{equation}
\textstyle	\beta\!=\!\big(1-\sum\nolimits_{k=1}^C{\underline{p}}_k\big)/\big({\sum\nolimits_{k=1}^C\Delta p_k}\big).
\label{Eq: beta}
\end{equation}
Here, $\Delta p_k$ denotes the $k$-th element of the interval length vector $\Delta\boldsymbol{p}\!=\!\overline{\boldsymbol{p}}\!-\!\underline{\boldsymbol{p}}$.
\subsection{Credal Student Design}
\label{subsec: CredalStudent}
\subsubsection{Architecture} Our proposed CREDIT merely modifies the final classification layer and is compatible with any NN backbone. 
Specifically, it replaces standard $C$ output neurons with $2C\!+\!1$ nodes to predict a vector $\boldsymbol{v}\!:=\!(\boldsymbol{p}_S^{*} \!\in\! \mathbb{R}^C, \Delta\boldsymbol{p}_S \!\in\! \mathbb{R}^C, \beta_S \!\in\! \mathbb{R})$, each component corresponding to the intersection probability, the interval length vector, and the weight factor, respectively. 
Let $\boldsymbol{z}_S\!\in\! \mathbb{R}^{2C+1}$ be the final layer logits outputs; then, $\boldsymbol{v}$ is computed as follows:
\begin{equation}
\begin{aligned}
\textstyle &\boldsymbol{p}_S^{*} \!=\! \operatorname*{softmax}({\boldsymbol{z}_S}_{1:C}); \Delta\boldsymbol{p}_S \!=\! \operatorname*{sigmoid}({\boldsymbol{z}_S}_{C+1:2C}); \\
\textstyle & \beta_S \!=\!\operatorname*{sigmoid}({z_S}_{2C+1})
\label{Eq: CredalStudentPrediction}
\end{aligned}.
\end{equation}
This ensures that $\boldsymbol{p}_S^{*}$ is normalized, that each $k$-th interval length $\Delta{p}_{S,k}\!:=\!{\overline{p}}_{S,k}\!-\!{\underline{p}}_{S,k}\!\in\![0, 1]$, and that $\beta_S\!\in\![0, 1]$. 
From \cref{Eq: intersection}, it is evident that 
the probability interval vector $[{\underline{\boldsymbol{p}}}_S, {\overline{\boldsymbol{p}}}_S]$
can be reconstructed from the elements of $\boldsymbol{v}$. Namely, its $k$-th element, $[{\underline{p}}_{S,k}, {\overline{p}}_{S,k}]$, can be computed as
\begin{equation}
\begin{aligned}
{\underline{p}}_{S,k} \!=\!{p}_{S,k}^{*} \!-\! \beta_S\Delta{p}_{S,k}, \ {\overline{p}}_{S,k} \!=\! {p}_{S,k}^{*} \!+\! (1\!-\!\beta_S)\Delta{p}_{S,k}
\label{Eq: PIreconstruct}
\end{aligned}.
\end{equation}
To guarantee that any ${\underline{p}}_{S,k}$ and ${\overline{p}}_{S,k}$ fall in $[0, 1]$, our credal ensemble distillation enforces:
\begin{equation}
\begin{aligned}
{\underline{p}}_{S,k} \!\gets\! \operatorname*{max}\{{\underline{p}}_{S,k}, 0\}, {\overline{p}}_{S,k} \!\gets\!\operatorname*{min}\{{\overline{p}}_{S,k}, 1\}
\label{Eq:ValidPI}
\end{aligned}.
\end{equation}
As a result, it can be proved that ${\overline{p}}_{S,k}\!-\!{\underline{p}}_{S,k}\!=\!\Delta{p}_{S,k} \!\in\! [0, 1]$ guarantees a valid probability interval for each class and that
\begin{equation}
	\begin{aligned}
		\!\!\textstyle\sum\nolimits_{k=1}^C{\underline{p}}_{S,k} \!\!=\!&\textstyle \sum\nolimits_{k=1}^C{p}_{S,k}^{*} \!\!-\! \beta_S\Delta{p}_{S,k}\! \leq\!\sum\nolimits_{k=1}^C{p}_{S,k}^{*} \!\!=\! 1 \\
		\!\!\leq & \textstyle \!\sum\nolimits_{k=1}^C{p}_{S,k}^{*} \!\!+\! (1\!-\!\beta_S)\Delta{p}_{S,k}\!\!=\!\! \sum\nolimits_{k=1}^C{\overline{p}}_{S,k}
	\end{aligned}
\end{equation}
satisfy the condition in \cref{Eq: CredalPIs} for defining a valid credal set for UQ.
$\boldsymbol{p}_S^{*}$ is then employed to predict a unique class.

\subsubsection{Uncertainty Quantification} To estimate both AU and EU from a credal set, one can use several measures, 
including the generalized entropy \citep{abellan2006disaggregated} and the generalized Hartley measure \citep{abellan2000non}. Due to its broad applicability and computational efficiency \citep{wang2024CredalEnsembles, wang2025credalwrapper}, this study adopts the generalized entropy measure. In this framework, an upper and a lower Shannon entropy, $\overline{H}(\mathbb{Q}_S)$ and $\underline{H}(\mathbb{Q}_S)$, are used to quantify total uncertainty (TU) and AU, respectively \citep{hullermeier2021aleatoric}. EU can then be estimated via $\overline{H}(\mathbb{Q}_S)\!-\!\underline{H}(\mathbb{Q}_S)$. Calculating $\overline{H}(\mathbb{Q}_S)$ here requires solving the following optimization problem:
\begin{equation}
\begin{aligned}
&\textstyle\overline{H}(\mathbb{Q}_S) \!=\!\operatorname*{maximize}\!\sum\nolimits_{k=1}^{C}\!-p_k\!\cdot\!\log p_k\\ 
&\text{s.t.}  \textstyle  \sum\nolimits_{k=1}^C\!p_k\!=\!1 \ \text{and} \ p_k \!\in\! [{\underline{p}}_{S,k}, {\overline{p}}_{S,k}]\ \forall k\!=\!1,...,C \ \
\label{Eq: CreUncertaintyImplementation}
\end{aligned},
\end{equation}
which searches for the maximum entropy value of any possible probability vector $\boldsymbol{p}$ within $\mathbb{Q}_S$. The computation of $\underline{H}(\mathbb{Q}_S)$, for which the $\operatorname*{maximize}$ is replaced by $\operatorname*{minimize}$, returns the minimal entropy. Standard solvers, such as the SciPy optimization package \citep{2020SciPy-NMeth}, can be used to efficiently solve these problems. Moreover, empirical evidence \citep{wang2024CredalEnsembles, wang2025credalwrapper} indicates that the computational overhead of \cref{Eq: CreUncertaintyImplementation} is marginal, particularly when $C\leq10$.

\subsection{Distillation Strategy}
\label{subsec: DistillationStrategy}
Generalizing cross-entropy (CE) loss, which corresponds to the Kullback-Leibler divergence, to the task of learning a credal set (defined by lower and upper probabilities) from a credal label remains an open research problem \citep{soubaras2011towards, lienen2023conformal, wang2024CredalEnsembles}. In this context, we propose minimizing the following loss, $\mathcal{L}_{\text{ced}}$, to distill a CREDIT student from the DE teacher:
\begin{equation}
\begin{aligned}
{N}^{-1}\!\sum\nolimits_{n=1}^{N}&\Big(\!\textstyle\sum\nolimits_{k=1}^C\!-{p^{*^n}_k}\!\log {p_{S,k}^{*^n}} \!+\! \\
&\textstyle\sum\nolimits_{k=1}^C\!{(\!\Delta p_k^{n}\!-\!\Delta p_{S,k}^{n})}^2 
\!+\!{(\beta^{n}\!-\!\beta_S^{n})}^2\!\Big)
\label{Eq: CEDloss}
\end{aligned}.
\end{equation}
Here, the superscript $n$ indicates the index of the $N$ training samples, and ${\underline{p}}_{S,k}^n$.
The first item of $\mathcal{L}_{\text{ced}}$ is the CE loss between the intersection probabilities of the DE ($\boldsymbol{p}^*$) and the CREDIT ($\boldsymbol{p}_S^*$). 
This enables CREDIT to retain the ensemble’s predictive performance—i.e., making a unique class prediction—since the intersection probability serves as the most representative single estimate for approximating probabilistic interval systems \citep{cuzzolin2022intersection}.
The second and third items correspond to classical mean-squared error losses that guide the student in learning the interval length $\Delta\boldsymbol{p}_S$ and the weight factor $\beta_S$ from the DE teacher, i.e., capturing the imprecision of the probability interval system (credal set) implied by the DE teacher's label. Following training, the probability interval systems in CREDIT are reconstructed via \cref{Eq: PIreconstruct}. Semantically, these last two items are equivalent to
\begin{equation}
\begin{aligned}
\textstyle\!\sum\nolimits_{k=1}^C\!{({\overline{p}}_{k}^{n}-{\overline{p}}_{S,k}^{n})}^2 \!+\! \sum\nolimits_{k=1}^C\!{({{\underline{p}}_{k}^{n}-\underline{p}}_{S,k}^{n})}^2
\label{Eq: CEDlossEqual}
\end{aligned},\nonumber
\end{equation}
where ${\overline{p}}_{S,k}^{n}$ and ${\underline{p}}_{k}^{n}$ can be calculated from ${p_{S,k}^*}^n$, $\Delta p_{S,k}^{n}$, and $\beta_S^{n}$ using \cref{Eq: PIreconstruct}.

Temperature scaling applied to the CE loss has been shown to enhance distillation performance \citep{hinton2015distilling}. This technique is also compatible with our CED framework, and the complete training procedure incorporating temperature scaling is outlined in Algorithm \ref{alg:CEDtraining}.
\begin{algorithm}[!htbp]
\caption{CED with Temperature Scaling}
\label{alg:CEDtraining}
\textbf{Input}: trained DE teacher (${\{\text{SNN}_m}\}_{m=1}^M$), temperature $T$\\
\textbf{Output}: credal student ($\text{CREDIT}$)
\begin{algorithmic}[1] 
\WHILE{training}
\STATE compute scaled logits of each $\text{SNN}_m$ given input $\boldsymbol{x}$: \\
$\boldsymbol{z}_m\!=\!\text{SNN}_m(\boldsymbol{x})/T$
\STATE calculate the predicted probability of each $\text{SNN}_m$: \\
$\boldsymbol{p}_m\!=\!\operatorname*{softmax}(\boldsymbol{z}_m)$
\STATE extract $\boldsymbol{p}^{*}$, $\Delta\boldsymbol{p}$, and $\beta$ using \cref{Eq: ExtractPI,Eq: intersection,Eq: beta}
\STATE compute scaled logits of the student given input $\boldsymbol{x}$: \\
$\boldsymbol{z}_S\!=\!({\boldsymbol{z}}_{1:C}/T, {\boldsymbol{z}}_{C+1:2C+1})$ where $\boldsymbol{z}\!=\ $\text{CREDIT}$(\boldsymbol{x})$
\STATE calculate $\boldsymbol{p}^{*}_S$, $\Delta\boldsymbol{p}_S$, and $\beta_S$ using \cref{Eq: CredalStudentPrediction}
\STATE minimize $T^2\!\cdot\!\mathcal{L}_{\text{ced}}$ in \cref{Eq: CEDloss}
\ENDWHILE
\end{algorithmic}
\end{algorithm}

\section{Experimental Validation}
\label{sec: Experiment}
To assess the UQ performance of our CED framework, empirical validations are conducted on standard OOD detection benchmarks \citep{mukhoti2023deep, mucsanyi2024benchmarking}. OOD detection is formulated as a binary classification task. ID and OOD samples are labeled as 0s and 1s, and the model's uncertainty estimate for each sample serves as its prediction. 
In this context, models are expected to assign higher EU/TU scores to OOD samples than to ID inputs. Thus, improved OOD detection performance serves as an indicator of better UQ. 
To quantify OOD detection performance, we report AUROC (Area Under the Receiver Operating Characteristic curve) and AUPRC (Area Under the Precision-Recall Curve) scores, where higher values reflect enhanced UQ performance.

\textbf{Setups\ } The empirical evaluation employs various dataset pairs (ID vs. OOD data), including CIFAR10 \citep{cifar10} vs. SVHN \citep{hendrycks2021natural} and CIFAR10 vs. CIFAR10-C \citep{hendrycks2019robustness}. The CIFAR10-C dataset consists of instances that apply 15 types of corruptions to the CIFAR10 test sets, respectively, with five severity levels per corruption type. 
\begin{table*}[t]
\centering
\scriptsize
\setlength\tabcolsep{5pt}
\begin{tabular}{@{}cl|cc|cccc|cccc@{}}
\toprule
                          &      & \multicolumn{2}{c|}{ID Performance}                    & \multicolumn{4}{c|}{OOD Detection (CIFAR10 vs. SVHN)}                                & \multicolumn{4}{c}{OOD Detection on (CIFAR10 vs. CIFAR10-C)}                           \\ \cmidrule(l){3-12} 
                          &      & \multicolumn{1}{c|}{Test}       & \multirow{2}{*}{ECE} & \multicolumn{2}{c|}{AUROC}                   & \multicolumn{2}{c|}{AUPRC} & \multicolumn{2}{c|}{AUROC}                   & \multicolumn{2}{c}{AUPRC} \\
                          &      & \multicolumn{1}{c|}{Accuracy}   &                      & Using EU   & \multicolumn{1}{c|}{Using TU}   & Using EU     & Using TU    & Using EU   & \multicolumn{1}{c|}{Using TU}   & Using EU     & Using TU   \\ \midrule \midrule
\multirow{6}{*}{VGG16}    & DE   & \multicolumn{1}{c|}{\textbf{93.52±0.07}} & \textbf{1.46±0.13}            & 89.99±0.79 & \multicolumn{1}{c|}{\underline{91.53±0.72}} & \underline{93.78±0.67}   & \underline{95.09±0.49}  & 93.18±1.99 & \multicolumn{1}{c|}{\fbox{\textbf{96.51±1.70}}} & \underline{89.41±4.07}   & \fbox{\textbf{95.42±2.07}} \\ \cmidrule(l){2-12} 
                          & SNN  & \multicolumn{1}{c|}{91.79±0.11} & 6.39±0.15            & /          & \multicolumn{1}{c|}{89.44±1.78} & /            & 93.71±1.24  & /          & \multicolumn{1}{c|}{93.90±2.41} & /            & 91.68±3.48 \\
                          & \textbf{CED}  & \multicolumn{1}{c|}{\underline{92.23±0.17}} & 6.71±0.18            & \fbox{\textbf{93.56±2.17}} & \multicolumn{1}{c|}{\textbf{92.51±1.96}} & \fbox{\textbf{96.09±1.72}}   & \textbf{95.21±1.52}  & \fbox{\textbf{96.51±1.81}} & \multicolumn{1}{c|}{\underline{95.56±1.75}} & \textbf{95.09±2.36}   & \underline{93.58±2.44} \\
                          & ED   & \multicolumn{1}{c|}{92.18±0.16} & 6.85±0.16            & /          & \multicolumn{1}{c|}{91.07±1.27} & /            & 94.51±0.89  & /          & \multicolumn{1}{c|}{94.71±2.20} & /            & 92.72±2.94 \\
                          & EDD* & \multicolumn{1}{c|}{91.13±0.18} & \underline{3.84±0.25}            & \underline{90.94±2.41} & \multicolumn{1}{c|}{90.96±2.66} & 93.66±1.72   & 93.78±2.11  & \underline{93.83±1.88} & \multicolumn{1}{c|}{95.45±2.10} & 87.91±4.32   & 92.11±3.65 \\ \cmidrule(l){2-12} 
                          & MCDO & \multicolumn{1}{c|}{91.95±0.13}           & 6.15±0.12             & 51.42±0.46 & \multicolumn{1}{c|}{89.12±1.63} & 74.72±0.42   & 93.64±1.17  & 51.32±0.50 & \multicolumn{1}{c|}{94.74±2.40} & 56.58 ± 1.92 & 93.12±3.10 \\ \midrule \midrule
\multirow{5}{*}{ResNet50} & DE   & \multicolumn{1}{c|}{\textbf{93.40±0.12}} & \textbf{1.32±0.16}            & \underline{89.50±1.05} & \multicolumn{1}{c|}{\textbf{94.89±0.50}} & 92.22±1.19   & \textbf{97.32±0.33}  & 87.78±2.28 & \multicolumn{1}{c|}{94.08±3.48} & 78.92±3.67   & \underline{93.08±3.92} \\ \cmidrule(l){2-12} 
                          & SNN  & \multicolumn{1}{c|}{91.60±0.38} & \underline{5.81±0.27}            & /          & \multicolumn{1}{c|}{93.55±0.96} & /            & 96.54±0.65  & /          & \multicolumn{1}{c|}{93.24±3.46} & /            & 91.64±4.27 \\
                          & \textbf{CED}  & \multicolumn{1}{c|}{91.77±0.74} & 6.34±0.59            & \fbox{\textbf{96.69±1.14}} & \multicolumn{1}{c|}{\underline{94.80±1.07}} & \fbox{\textbf{98.44±0.64}}   & \underline{97.12±0.67}  & \fbox{\textbf{96.80±2.81}} & \multicolumn{1}{c|}{\textbf{95.23±2.74}} & \fbox{\textbf{96.09±4.14}}   & \textbf{93.78±3.72} \\
                          & ED   & \multicolumn{1}{c|}{\underline{92.02±0.22}} & 6.64±0.24            & /          & \multicolumn{1}{c|}{94.02±0.55} & /            & 96.50±0.43  & /          & \multicolumn{1}{c|}{\underline{94.09±2.81}} & /            & 92.22±3.69 \\
                          & EDD* & \multicolumn{1}{c|}{80.38±6.10} & 10.79±8.12           & 86.80±8.08 & \multicolumn{1}{c|}{88.75±5.93} & \underline{93.27±4.74}   & 93.61±3.94  & \underline{89.48±9.52} & \multicolumn{1}{c|}{91.04±6.75} & \underline{86.30±12.72}  & 86.76±9.45 \\ \bottomrule
\end{tabular}
\caption{ID performance (test accuracy and ECE) and OOD detection performance comparison across methods on benchmarks (CIFAR10 vs. SVHN/CIFAR10-C) involving VGG16 and ResNet50 (in a pre-trained model setting) as backbones. Results (all in \%) are averaged from 15 runs. In terms of OOD detection, ranking legend for consistent use of uncertainty estimates (either EU or TU): \textbf{Best}, \underline{Second best}. The highest scores across both EU and TU are indicated with {\scriptsize\fbox{\phantom{best}}}.} 
\label{Table: MainComparison}
\end{table*}

\begin{figure*}[!htbp]
\centering
\includegraphics[width=0.925\textwidth]{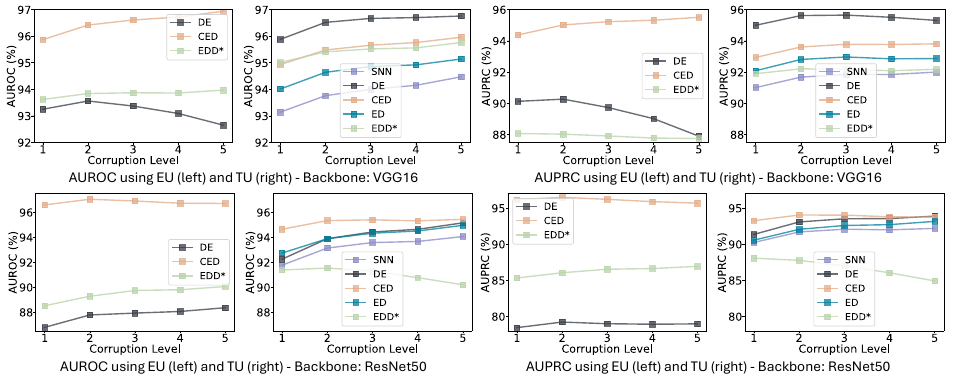}
\caption{OOD detection (CIFAR10 vs. CIFAR10-C) comparison over increased corruption levels on various backbones.}
\label{Fig: RES18VGG16RES5010C}
\end{figure*}

The experiment begins by training 15 SNNs with different random initializations using the ID training sets. Choosing 15 runs aims to improve the statistical significance of the results. DEs are constructed by five SNN models ($M = 5$). To prevent distilled models from relying on one particular DE teacher, we construct 15 DEs by randomly selecting 15 distinct subsets of size 5 from the full SNN index set $\{1, ..., 15\}$, ensuring that no ensemble shares an identical composition. Subsequently, three types of student models—ED, EDD, and our CED—are distilled from these DEs, yielding 15 student models per category. All training configurations, including batch size, number of epochs, optimizer, learning rate scheduler, and temperature scaling ($T = 2.5$) as recommended by \citep{hinton2015distilling}, are kept consistent within each student class to ensure fair comparison. Additionally, we include a variant of EDD, denoted as EDD*, which adopts the cyclic learning rate policy, temperature scaling ($T = 10$), and temperature annealing, following the recipe in the original study \citep{malinin2019ensemble}. In addition, we include the well-known Monte Carlo Dropout (MCDO) \citep{gal2016dropout} with 10 forward passes during inference as an additional baseline.

As the main experiment, all models are implemented on the established VGG16 architecture \citep{simonyan2015very}, and trained from scratch using the CIFAR10 dataset. As an ablation study, we also evaluate the UQ performance of our CED method in a pre-trained model setting. All models are trained on the CIFAR10 dataset using pre-trained ResNet-50 backbones \citep{he2016deep}, following the same training pipeline as in the main experiment. To accommodate the pre-trained models, all input images are resized to $(224, 224, 3)$. Because no pretrained models are available, we do not report ResNet50-based MCDO results.

\textbf{Results \ }
Table \ref{Table: MainComparison} presents UQ evaluations on EU and TU
on the OOD benchmarks (CIFAR10 vs. SVHN/CIFAR10-C). For CIFAR10-C as OOD data, the reported scores are averaged over 15 corruption types and 5 severity levels. The EU and TU estimates of DE are computed from \cref{Eq: AU_BNN} in a standard manner. Figure \ref{Fig: RES18VGG16RES5010C} shows the OOD detection comparison on CIFAR10 vs. CIFAR10-C against increased corruption intensity. Evaluation results for EDD, trained with the same configurations as ED and CED for a fair comparison, are excluded due to its substantially lower prediction accuracy, as shown in \tablename~\ref{Table: PoorEDDOnly}.
\begin{table}[!htbp]
\centering
\scriptsize
\setlength\tabcolsep{5pt}
\begin{tabular}{@{}lcc|cc@{}}
\toprule
     & ACC [VGG16]         & ECE  [VGG16]                & ACC   [ResNet50]        & ECE  [ResNet50]        \\ \midrule
EDD  & 74.56±2.02 & 5.51±0.57  & 61.15±8.47  & 8.40±2.06  \\ \bottomrule
\end{tabular}
\caption{Test ACC and ECE of EDD on CIFAR10 test data across distinct architectures. All results are in \%.}
\label{Table: PoorEDDOnly}
\end{table}
\begin{figure}[!htbp]
\centering
\includegraphics[width=0.9\columnwidth]{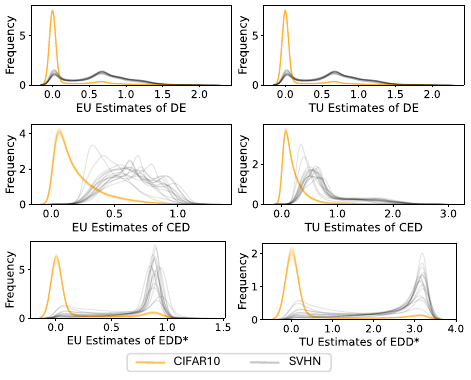}
\includegraphics[width=0.9\columnwidth]{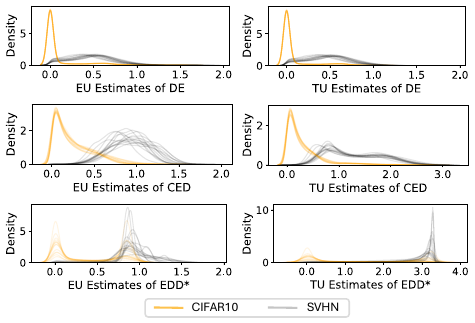}
\caption{Distributions of EU and TU estimates across models on the VGG16 (top) and ResNet50 (bottom). 15 runs.}
\label{Fig: UncertaintyPlots}
\end{figure}

These OOD detection results in \tablename~\ref{Table: MainComparison} show that our CED framework \emph{significantly} and \emph{consistently} improves EU estimation over baseline methods, as reflected by enhanced OOD detection performance across diverse dataset pairs and backbone architectures. For TU estimation, CED achieves superior or comparable performance over baselines, ranking among the top two in most cases. Furthermore, the comparison of OOD detection scores using both EU and TU in Table \ref{Table: MainComparison} and Figure \ref{Fig: RES18VGG16RES5010C} shows that CED's EU estimates \emph{yield the best performance in most cases}. This highlights the importance of improving EU quantification for reliable OOD detection. As a summary, these empirical results establish CED as a principled alternative distillation framework for UQ. In this setting, EU estimation with MCDO becomes unreliable, likely due to limited model diversity.

In addition, \tablename~\ref{Table: MainComparison} also presents test accuracy and expected calibration error (ECE) \citep{guo2017calibration} on the CIFAR10 test set for various models. A lower ECE value signifies a closer alignment between the model’s confidence scores and the true probabilities of the events \citep{guo2017calibration, nixon2019measuring}. The results indicate that distillation enhances the predictive accuracy of individual SNNs, and our CED approach achieves performance comparable to baseline distillation methods. Note that the ECE metric used here is designed for single-probability predictions, and a principled extension of ECE to credal-set predictions is needed for a fair comparison \citep{wang2024CredalEnsembles}.

\textbf{Qualitative Evaluation \ }
Figure \ref{Fig: UncertaintyPlots} displays kernel density plots of EU and TU estimates on CIFAR10 (ID) and SVHN (OOD) samples for DE, CED, and EDD*, across different backbone architectures. While the EU and TU estimates are not directly comparable across methods due to differing uncertainty representations, CED consistently exhibits substantially higher EU and TU values for OOD samples relative to ID instances, as qualitatively observed. 

It is worth noting that in the VGG16 case in Figure \ref{Fig: UncertaintyPlots}, although EDD* yields more distinct density peaks in EU and TU estimates for OOD samples, the uncertainty distributions for ID samples also exhibit considerable density near the OOD peaks. This overlap aligns with the lower OOD detection performance of EDD* compared to DE and CED, as reported in Table \ref{Table: MainComparison}. The peaks for EDD* on ResNet50 stem from its low test accuracy (\tablename~\ref{Table: MainComparison}), which results in more misclassified ID samples with high uncertainty.

\textbf{Ablation Study on Teacher Ensemble Size \ }
This experiment assesses the UQ performance of our CED across varying ensemble sizes of the teacher model. Using the same training procedure as in the main experiment, we consider DE teachers with $M \in \{5, 15, 25, 30\}$ for distillation.
\begin{figure*}[!htbp]
\centering
\includegraphics[width=0.9\textwidth]{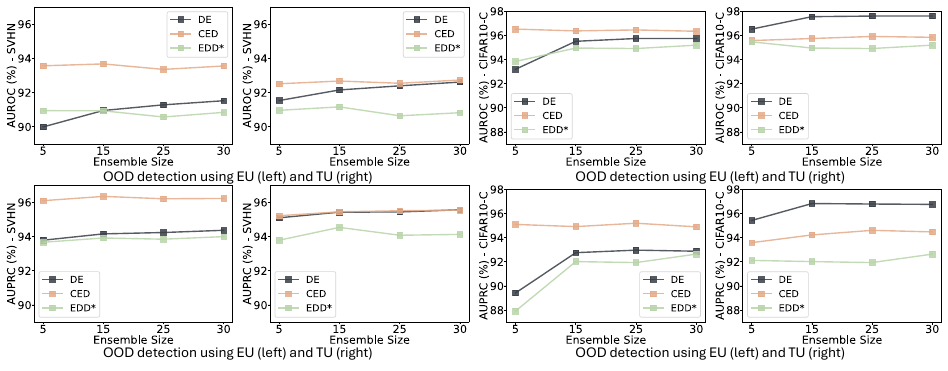}
\caption{OOD detection performance with increasing ensemble sizes of the DE teacher. Left: CIFAR10 vs. SVHN. Right: CIFAR10 vs. CIFAR10-C. Backbone: VGG16.}
\label{Fig: VGG16EnsembleFull}
\end{figure*}

Figure \ref{Fig: VGG16EnsembleFull} presents the OOD detection performance involving the VGG16 backbone for CIFAR10 vs. SVHN and CIFAR10 vs. CIFAR10-C, respectively. The results reveal the following observations: Unlike the DE teacher, where increasing the ensemble size consistently improves UQ performance, no such clear trend is observed for either CED or EDD*. The consistently strong OOD detection scores with EU, along with comparable results using TU, highlight the high potential of our CED approach for UQ, particularly given the significantly lower inference complexity compared to DE with larger ensemble sizes. 

\begin{figure}[!htbp]
\centering
\includegraphics[width=0.925\columnwidth]{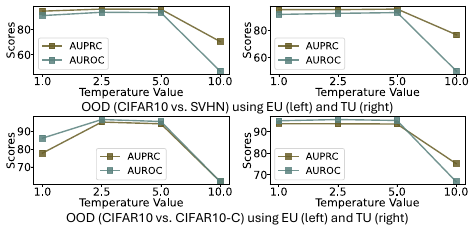}
\caption{OOD detection performance over increased temperature $T$ values. Backbone: VGG16.}
\label{Fig: VGG16Temp}
\end{figure}

\textbf{Ablation Study on Effect of Temperature Scaling \ } 
This experiment empirically investigates the effect of various 
temperature scaling on the UQ performance of our CED. Using the same training procedure as in the main experiment, we consider $T \in \{1, 2.5, 5, 10\}$ (recommended by \citep{hinton2015distilling}) for distillation. Figure \ref{Fig: VGG16Temp} presents the OOD detection scores on CIFAR10 vs. SVHN/CIFAR10-C using the VGG16 backbone across various temperature scaling values. The results indicate that temperature scaling improves the UQ performance of our CED method, although excessively high values (i.e., $T = 10.0$) degrade performance. Among the tested values, $T = 2.5$ consistently yields the best results, aligning with the findings of \citet{hinton2015distilling}. 

\textbf{Complexity \ } 
Table \ref{Table: InferenceComparison} presents the inference times in seconds on the CIFAR10 test set using a single P100 GPU. 
\begin{table}[!htbp]
\centering
\scriptsize
\begin{tabular}{@{}l|ccc@{}}
\toprule
& DE             & CED                 & EDD*         \\ \midrule
Inference time   & 5$\times$(2.22±\scriptsize0.20) & 2.26±\scriptsize0.23 & 2.22±\scriptsize0.20 \\ \midrule
Training time   & 5$\times$(130.07±\scriptsize0.24) & 659.52±\scriptsize11.82 & 684.54±\scriptsize5.05 \\
\bottomrule
\end{tabular}
\caption{Inference time (s) comparison on CIFAR10 test set.}
\label{Table: InferenceComparison}
\end{table}
While CED introduces a slight increase in inference complexity compared to other distillation methods—due to additional output layer nodes—it remains significantly more efficient than DE. In addition, the training time per epoch in seconds using a single
P100 GPU in Table \ref{Table: InferenceComparison} shows that training CED is simpler than training EDD*, as CED does not require a sophisticated learning rate scheduler or temperature annealing.

Additional experimental details, ablation studies on the ResNet18 backbone, and a case study on real-world medical image classification are provided in the Appendix.
\section{Conclusion and Future Work}
\label{Sec: conclude}
This work presented \emph{credal ensemble distillation} (CED), a novel framework that compresses a deep ensemble (DE) teacher into a single model, \emph{CREDIT}, for classification tasks. Instead of a single softmax probability distribution, CREDIT is capable of predicting class-wise probability intervals that define a credal set for uncertainty quantification. Empirical results on out-of-distribution detection benchmarks demonstrated that CED achieves superior or comparable uncertainty estimation compared to several existing ED, EDD and DE baselines, while substantially reducing inference overhead compared to DE.  We believe that these promising results could position our credal ensemble distillation as a principled and scalable alternative for uncertainty quantification in deep neural classifiers.

One direction for future work is to enhance the scalability of CED, enabling its application to classification tasks involving a significantly larger number of classes (e.g., 100 or 1000). A key challenge in the current CED setting is that the softmax activation used in DE teacher produces extremely small probability values (near zero) for most classes, which could potentially destabilize the regression component of the distillation loss during training and undermine the robustness of CED’s uncertainty quantification. Another future goal is to integrate calibration considerations into the design of the distillation strategy, with the aim of achieving comparable or better calibration performance than the DE teacher.

\section*{Acknowledgments}
This work has received funding from the European Horizon 2020 research and innovation program under the FET Open grant agreement No. 964505 (E-pi).

\bibliography{aaai2026}
\newpage
\appendix
\setcounter{secnumdepth}{2}
\section{Experiment Details}
\label{App: ExperimentDetails}
\subsection{Training Configuration}
\subsubsection{Training Details} \ In terms of the \emph{main experiment,} all models, including standard neural networks (SNNs), ensemble distillation (ED), ensemble distribution distillation (EDD), our credal ensemble distillation (CED), and Monte Carlo Dropout (MCDO) with a dropout rate of 0.1 are implemented on the established VGG16 and ResNet18 architectures using the CIFAR10 dataset. Standard data augmentation is uniformly implemented across all methodologies to enhance the training performance quality of training data and training performance. The training batch size is set as 128. The standard data split is applied. The Adam optimizer is applied with a learning rate scheduler, initialized at 0.001, and subjected to a 0.1 reduction at epoch 80. All models are trained for 100 epochs. Random initialization is applied for all models.
Additionally, we also train a variant of EDD for 100 epochs, denoted as EDD*, which adopts the cyclic learning rate policy (with cycle length 60), temperature scaling ($T = 10$), and temperature annealing, following the recipe in the original study \citep{malinin2019ensemble}. All other training configurations, such as data splitting and augmentation, are kept consistent with those used for the other models. 

Regarding \emph{ablation study of pre-trained model utilization}, a single Nvidia A100-SXM4-80GB GPU is used as the device. The input shape of the networks is (224, 224, 3). The Adam optimizer is employed, with a learning rate scheduler set at 0.001 and reduced to 0.0001 during the final 5 training epochs. All models are trained for 25 epochs under different random seeds, using the CIFAR10 dataset. Temperature scaling $T=2.5$ is applied to ED, EDD, and CED, the same as the main experiment. Regarding EDD*, the cycle length for the cyclic learning rate policy is set as 15. Temperature scaling ($T = 10$) and temperature annealing are applied.  All other training configurations, such as data splitting, are kept consistent with those used for the other models. 

The main experimental implementation codes are provided in the supplementary files.

\subsubsection{EDD Training Strategy} Distilling a DE consisting of $M$ trained SNNs into an EDD-Net can be achieved by minimizing the following likelihood loss \citep{malinin2019ensemble}: 
\begin{equation}
\begin{aligned}
-\frac{1}{N}\sum\nolimits_{n=1}^{N}\Big(&\ln\Gamma(\alpha_0^{n})-\sum\nolimits_{k=1}^{C}\ln\Gamma(\alpha_k^{n})\\
&+{M}^{-1}\sum\nolimits_{m=1}^{M}\sum\nolimits_{k=1}^{C}(\alpha_k^{n}-1)\ln p_{m,k}^{n}\Big)  
    \label{Eq: EDDLoss},
\end{aligned}
\end{equation}
where $\Gamma$ is the Gamma function and the superscript $n$ corresponds to the index of the training sample. For example, $p_{m,k}^{n}$ denotes the $k$-th probability component of the soft label produced by the $m$-th ensemble member.
\subsection{OOD Detection Process and ECE Calculation}
\subsubsection{OOD Detection Process}
In this paper, the out-of-distribution (OOD) detection process is treated as a binary classification. We label ID and OOD samples as 0s and 1s, respectively. The model’s uncertainty estimation, using the epistemic uncertainty (EU) or total uncertainty (TU), for each sample, is the `prediction' for the detection. In terms of performance indicators, the applied AUROC quantifies the rates of true and false positives. The AUPRC evaluates precision and recall trade-offs, providing valuable insights into the model's effectiveness across different confidence levels. The OOD detection process is summarized in Algorithm \ref{alg: ood-algorithm}.
\begin{algorithm}[!hbtp]
\begin{algorithmic}
\STATE\textbf{Input:} Uncertainty estimates for ID and OOD samples, namely $\boldsymbol{u}_{\text{ID}}$, $\boldsymbol{u}_{\text{OOD}}$
\STATE \textbf{Output:} AUROC and AUPRC scores
\STATE \textbf{1.} Set labels ($\boldsymbol{b}_{\text{ID}}$) as 0 for ID samples 
\STATE \quad $\boldsymbol{b}_{\text{ID}} \gets \text{zeros}(\text{shape of } \boldsymbol{u}_{\text{ID}})$
\STATE \textbf{2.} Set labels ($\boldsymbol{b}_{\text{OOD}}$) as 1 for OOD samples
\STATE \quad $\boldsymbol{b}_{\text{OOD}} \gets \text{ones}(\text{shape of } \boldsymbol{u}_{\text{OOD}})$
\STATE \textbf{3.} Concatenate labels for all samples
\STATE \quad $\boldsymbol{b} \gets \text{concatenate}(\boldsymbol{b}_{\text{ID}}, \boldsymbol{b}_{\text{OOD}})$
\STATE \textbf{4.} Concatenate uncertainty estimates as ``predictions'' 
\STATE \quad $\boldsymbol{u} \gets \text{concatenate}(\boldsymbol{u}_{\text{ID}}, \boldsymbol{u}_{\text{OOD}})$
\STATE \textbf{5.} Compute AUROC and AUPRC values 
\STATE \quad $\text{AUROC} \gets \text{roc\_auc\_score}(\boldsymbol{b}, \boldsymbol{u})$
\STATE \quad $\text{AUPRC} \gets \text{average\_precision\_score}(\boldsymbol{b}, \boldsymbol{u})$
\caption{OOD Detection Process} 
\label{alg: ood-algorithm}
\end{algorithmic}
\end{algorithm}

\subsubsection{ECE Evaluation} \ In the context of expected calibration error (ECE), a `well-calibrated' prediction is expected to have a confidence value of 80\% and be correct in approximately 80\% of the test cases. 

To calculate ECE, predictions are split into a predetermined number $Q$ of bins $B$ of equal confidence range. The ECE is then calculated by summing the absolute difference between the average accuracy and confidence within each bin \citep{mehrtens2023benchmarking}:
\begin{equation}
\text{ECE} \!:=\!\sum\nolimits_{g=1}^{G}\frac{|B_g|}{n}\bigg|\text{acc}(B_g)-\text{conf}(B_g)\bigg|
\label{Eq: ECE},
\end{equation}
where $|B_g|$ is the number of samples in the $g$-th bin and $n$ is the total number of samples.

\section{Ablation Studies on ResNet18 Backbone}
\subsection{Ablation Study for Main Experiments}
\begin{figure}[t]
\centering
\includegraphics[width=0.925\linewidth]{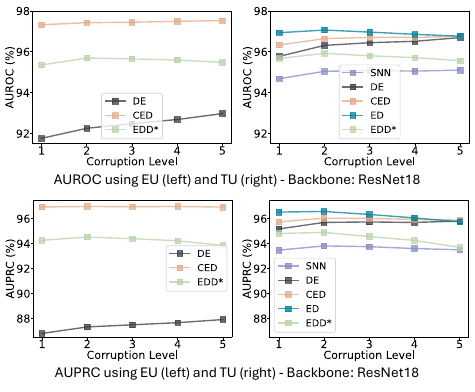}
\caption{OOD detection (CIFAR10 vs. CIFAR10-C) comparison over increased corruption levels on the ResNet18 backbone.}
\label{Fig: RES18}
\end{figure}

These OOD detection results in \tablename~\ref{Table: MainComparisonAblationStudy} show that our CED framework \emph{significantly} and \emph{consistently} improves EU estimation over baseline methods, as reflected by enhanced OOD detection performance across diverse dataset pairs and backbone architectures. For TU estimation, CED achieves superior or comparable performance over baselines, ranking among the top two in most cases. Furthermore, the comparison of OOD detection scores using both EU and TU in Table \ref{Table: MainComparisonAblationStudy} and Figure \ref{Fig: RES18} shows that CED's EU estimates \emph{yield the best performance in most cases}. This highlights the importance of improving EU quantification for reliable OOD detection. As a summary, these empirical results establish CED as a principled alternative distillation framework for UQ. In this setting, EU estimation with MCDO becomes unreliable, likely due to limited model diversity.
\begin{table}[!htbp]
\centering
\scriptsize
\begin{tabular}{@{}lcccc@{}}
\toprule
& \multicolumn{2}{c|}{SVHN (OOD)} & \multicolumn{2}{c}{CIFAR10-C (OOD)} \\
\cmidrule(l){2-5}
& AUROC & \multicolumn{1}{c|}{AUPRC} & AUROC & AUPRC \\
\midrule
\midrule
& \multicolumn{4}{c}{OOD Detection using EU} \\
\midrule
\multicolumn{1}{l|}{DE} & {87.63±0.57} & {91.99±0.49} & {92.43±1.91} & {87.91±4.32} \\
\midrule
\multicolumn{1}{l|}{CED} & \textbf{88.73±2.53} & \textbf{92.66±1.87} & \fbox{\textbf{97.44±1.35}} & \fbox{\textbf{96.95±1.80}} \\
\multicolumn{1}{l|}{EDD*} & \underline{87.77±1.82} & \underline{92.39±1.03} & \underline{95.55±1.91} & \underline{94.25±2.84} \\ \midrule
\multicolumn{1}{l|}{MCDO} & 50.76±0.32 & 76.52±0.34 & 50.83±0.42 & 60.00±1.16 \\
\midrule
\midrule
& \multicolumn{4}{c}{OOD Detection using TU} \\
\midrule
\multicolumn{1}{l|}{SNN} & 87.60±1.99 & 92.20±1.33 & 94.99±1.79 & 93.63±2.50 \\
\multicolumn{1}{l|}{DE} & 88.76±0.60 & \underline{93.09±0.52} & 96.34±1.71 & 95.62±2.11 \\
\midrule
\multicolumn{1}{l|}{CED} & \underline{88.81±1.71} & 93.02±1.03 & \underline{96.62±1.47} & \underline{95.92±1.95} \\
\multicolumn{1}{l|}{ED} & \fbox{\textbf{89.21±1.56}} & \fbox{\textbf{93.36±0.96}} & \textbf{96.91±1.43} & \textbf{96.24±2.08} \\
\multicolumn{1}{l|}{EDD*} & 87.45±2.00 & 91.50±1.43 & 95.73±1.74 & 94.44±2.55 \\ \midrule
\multicolumn{1}{l|}{MCDO} & 86.51±1.57 & 91.63±1.11 & 95.23±1.66 & 93.99±2.33 \\
\bottomrule
\end{tabular}
\caption{Performance comparison across methods on OOD detection benchmarks (CIFAR10 vs. SVHN/CIFAR10-C) involving the ResNet18 backbone. Results (all in \%) are averaged from 15 runs. Ranking legend for consistent use of uncertainty estimates (either EU or TU): \textbf{Best}, \underline{Second best}. Highest scores across EU and TU are indicated with {\scriptsize\fbox{\phantom{best}}}.} 
\label{Table: MainComparisonAblationStudy}
\end{table}
Table \ref{Table: PoorEDD} presents test accuracy (ACC) and expected calibration error (ECE) \citep{guo2017calibration} on the CIFAR10 test set for various models using the ResNet18 backbone. The results indicate that distillation (ED, EDD*, and CED) can enhance the predictive accuracy of individual SNNs, and our CED approach achieves performance comparable to baseline distillation methods. However, no single models match the calibration performance of the DE teacher. This motivates us to integrate calibration considerations into the design of the distillation strategy in future work to achieve comparable or better calibration performance than the DE teacher. Table \ref{Table: PoorEDD} also shows EDD's significantly lower and unreliable prediction accuracy. Therefore, its uncertainty quantification analysis is excluded.
\begin{table}[!htbp]
\centering
\scriptsize
\begin{tabular}{@{}lcc@{}}
\toprule
     & ACC            & ECE                  \\ \midrule
DE   & 93.37±0.12   & 1.84±0.14  \\ \midrule \midrule
SNN  &  92.03±0.15   & 5.88±0.13   \\
CED  &  92.93±0.09   & 4.82±0.08    \\
ED   &  92.87±0.15  & 4.89±0.17    \\
EDD* & 92.95±0.13   & 2.75±0.13   \\
EDD  & 85.29±1.29   & 6.80±0.17   \\ \midrule
MCDO & 92.33±0.17 &5.25±0.17 \\
\bottomrule
\end{tabular}
\caption{Test ACC and ECE on CIFAR10 test data across distinct models using various neural network settings. All results are in \%.}
\label{Table: PoorEDD}
\end{table}
\subsection{Ablation Study on Teacher Ensemble Size}
\begin{figure}[!htbp]
\centering
\includegraphics[width=0.95\columnwidth]{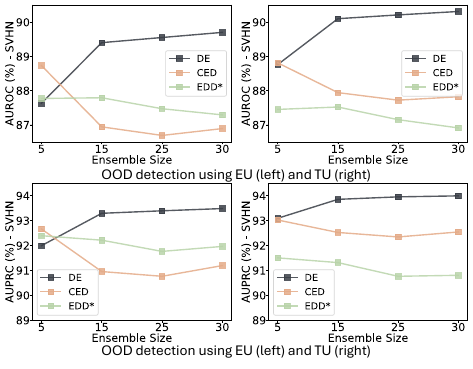}
\caption{OOD detection (CIFAR10 vs. SVHN) over increased ensemble sizes of DE teacher.}
\label{Fig: RES18EnsembleSVHN}
\end{figure}

\begin{figure}[!htbp]
\centering
\includegraphics[width=0.95\columnwidth]{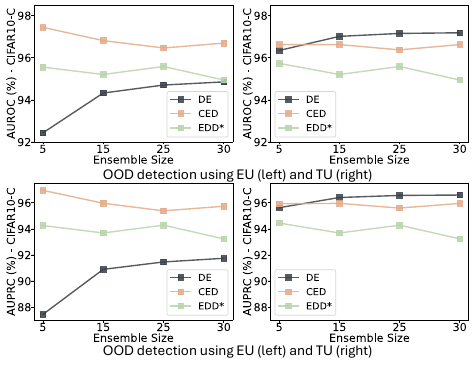}
\caption{OOD detection (CIFAR10 vs. CIFAR10-C) over increased ensemble sizes of DE teacher.}
\label{Fig: RES18EnsembleCIFAR10C}
\end{figure}

Figures \ref{Fig: RES18EnsembleSVHN} and \ref{Fig: RES18EnsembleCIFAR10C} present the OOD detection performance involving the ResNet18 backbone for CIFAR10 vs. SVHN and CIFAR10 vs. CIFAR10-C, respectively. The results reveal the following observations: Unlike the DE teacher, where increasing the ensemble size consistently improves UQ performance, no such clear trend is observed for either CED or EDD*. The consistently strong OOD detection scores with EU and TU in most cases highlight the high potential of our CED approach for UQ, particularly given the significantly lower inference complexity compared to DE with larger ensemble sizes.
\subsection{Ablation Study on Effect of Temperature Scaling}
\begin{figure}[!htbp]
\centering
\includegraphics[width=0.95\columnwidth]{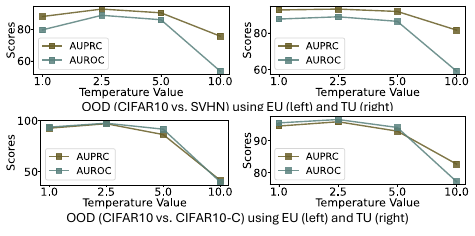}
\caption{OOD detection performance over increased temperature $T$ values. Backbone: ResNet18.}
\label{Fig: RES18Temp}
\end{figure}

Figure \ref{Fig: RES18Temp} presents the OOD detection scores on CIFAR10 vs. SVHN/CIFAR10-C across various temperature scaling values, showing that temperature scaling improves the UQ performance of the CED, although excessively high values (i.e., $T = 10.0$) degrade performance. Among tested values, $T = 2.5$ consistently yields the best results, aligning with the findings of \citet{hinton2015distilling}. 
\section{Case Study on Real-World Medical Image Binary Classification}
\subsection{Experiment Description}
\begin{figure}[!htbp]
\begin{center}
\includegraphics[width=0.95\linewidth]{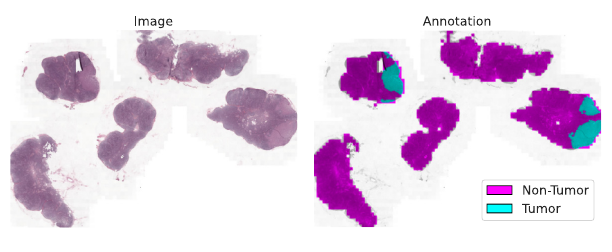}
\end{center}
\caption{An example of the whole slide image (referring the node 2 of patient 017  in the Camelyon17 dataset) with ground-truth annotations.}
\label{FIG: WSIimage}
\end{figure}

Following the work \citep{mehrtens2023benchmarking}, this case study employs the Camelyon17 \citep{bandi2018detection} histopathological dataset for binary \{Tumor, Non-Tumor\} classification, which contains lesion-level annotated whole slide images (WSIs) of breast lymph node tissue with metastatic regions. The Camelyon17 dataset comprises 50 WSIs collected from five medical centers in the Netherlands using three different scanners. An example of WSIs is shown in Figure \ref{FIG: WSIimage}.

To consider a strong domain shift setting \citep{mehrtens2023benchmarking}, the dataset is partitioned such that the out-of-distribution (OOD) set contains only images acquired with scanners absent from the in-distribution (ID) set, thereby introducing an additional technological shift in image acquisition. Specifically, centers 0, 1, and 3—equipped with 3DHistech scanners—constitute the ID set, while centers 2 and 4—using Philips and Hamamatsu scanners—form the OOD set.

In this case study, we first train the DE model ($M=5$) following the procedure in \citet{mehrtens2023benchmarking}, and then distill a CED from the trained DE.

Because the task is binary classification, the credal set is expressed by a single probability interval $[\underline{p}_S, \overline{p}_S]$
. Following recent work proposing more principled and efficient alternatives \citep{hullermeier2022quantification}, we adopt the following measures:
\begin{equation}
\text{EU} := \overline{p}_S - \underline{p}_S, \quad
\text{TU} := \min(1-\underline{p}_S, \overline{p}_S).
\label{Eq: uncertaintiesInterval}
\end{equation}
For a more detailed discussion of these uncertainty measures, including their advantages and limitations, we refer the reader to \citep{hullermeier2022quantification}.
\subsection{Evaluation and Results}
We evaluate the quality of the EU and TU estimates using accuracy–rejection (AR) curves in both ID and OOD settings. AR curves show how a model’s prediction accuracy changes as a function of the rejection rate in selective classification \citep{hullermeier2022quantification}. Given a batch of instances, those with higher uncertainty are rejected first, and the accuracy is computed on the remaining samples. We generate AR curves by rejecting samples based on either the model’s EU or TU estimates. A well-calibrated uncertainty measure yields a monotonically increasing AR curve, whereas random rejection leads to a largely flat curve \citep{hullermeier2022quantification}. We also report the area under the AR curve (AUARC) as a comparative metric, with higher AUARC values indicating better performance.

\begin{figure}[!htbp]
\begin{center}
\includegraphics[width=0.8\linewidth]{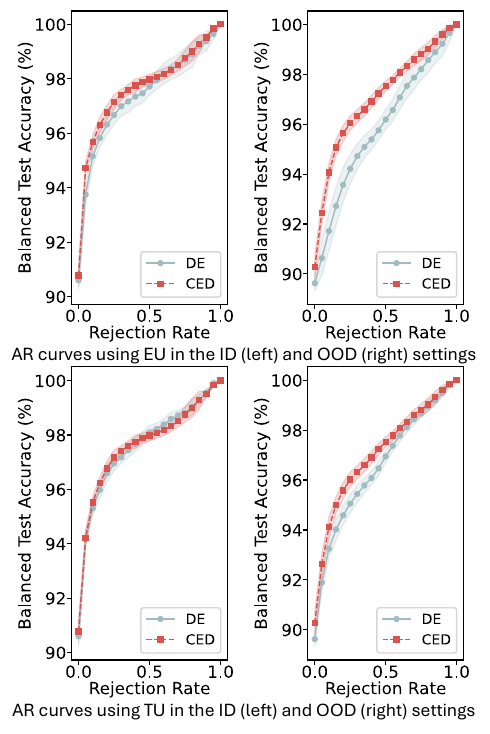}
\end{center}
\caption{Balanced test accuracy-rejection curves in different settings.}
\label{FIG: AR}
\end{figure}

\figurename~\ref{FIG: AR} shows the averaged balanced test accuracy-rejection curves in different settings over $5$ runs and \tablename~\ref{Tab: AUARC} reports the averaged AUARC scores. These results show that CED outperforms DE while also requiring less computation.

\begin{table}[!htbp]
\centering
\scriptsize
\begin{tabular}{lccc}
\toprule
Setting & Estimates & CED & DE \\
\midrule
\multirow{2}{*}{ID}
    & EU & 97.71 ± 0.20 & 97.43 ± 0.34 \\
    & TU & 97.67 ± 0.20 & 97.65 ± 0.22 \\
\midrule
\multirow{2}{*}{OOD}
    & EU & 97.12 ± 0.22 & 95.92 ± 0.44 \\
    & TU & 97.12 ± 0.22 & 96.61 ± 0.24 \\
\bottomrule
\end{tabular}
\caption{Averaged AUARC scores (\%) over 5 runs.}
\label{Tab: AUARC}
\end{table}

\end{document}